\documentclass[pmlr]{jmlr}
\usepackage[utf8]{inputenc}
\usepackage[english]{babel}
\usepackage{amsfonts}
\usepackage{algorithm}
\usepackage{graphicx}
\graphicspath{ {./images/} }
\usepackage{float}
\usepackage[load-configurations=version-1]{siunitx}

\usepackage{booktabs}
\usepackage{chemformula}

\title{Path Signature Area-Based Causal Discovery in Coupled Time Series}

\author{
\Name{Will Glad, MS} \Email{wglad1@alumni.jh.edu}\\
\addr Johns Hopkins University, Applied \& Computational Mathematics
\AND
\Name{Tom Woolf, PhD} \Email{twoolf@jhu.edu}\\
\addr Johns Hopkins University, Applied \& Computational Mathematics
}
\begin{document}
\maketitle

\begin{abstract}
Coupled dynamical systems are frequently observed in nature, but often not well understood in terms of their causal structure without additional domain knowledge about the system.  Especially when analyzing observational time series data of dynamical systems where it is not possible to conduct controlled experiments, for example time series of climate variables, it can be challenging to determine how features causally influence each other.  There are many techniques available to recover causal relationships from data, such as Granger causality, convergent cross mapping, and causal graph structure learning approaches such as PCMCI. Path signatures and their associated signed areas provide a new way to approach the analysis of causally linked dynamical systems, particularly in informing a model-free, data-driven approach to algorithmic causal discovery.  With this paper, we explore the use of path signatures in causal discovery and propose the application of confidence sequences to analyze the significance of the magnitude of the signed area between two variables. These confidence sequence regions converge with greater sampling length, and in conjunction with analyzing pairwise signed areas across time-shifted versions of the time series, can help identify the presence of lag/lead causal relationships. This approach provides a new way to define the confidence of a causal link existing between two time series, and ultimately may provide a framework for hypothesis testing to define whether one time series causes another.
\end{abstract}

\section{Introduction}
Path signatures developed from analysis of stochastic differential equations and have been brought to the forefront of modern methods by the group surrounding \cite{lyons2014rough}. There is a growing body of literature evaluating the use of the path signature as a feature extraction technique for performing time series classification and prediction, such as \cite{chevyrev2016primer}, \cite{lyons2014rough}, \cite{levin2016learning}, \cite{lemercier2020distribution}, \cite{signatory}, \cite{FERMANIAN2021107148} and \cite{morrill2021generalised}, to name only a few. 

Causal ideas and their statistical nuances have been considered by many researchers over many years, such as \cite{GRANGER1980329, robins2000marginal}, and \cite{pearl2009causality}. To our knowledge, path signatures have not been extensively studied in the context of causal inference, with \cite{giusti2020iterated} being the first attempt to explicitly study the signature's properties in causal identification for multivariate time series. 

In this paper, we focus specifically on the task of discovering causal relationships from observational time series data from coupled dynamical systems using the signature method, and in particular by evaluating the pairwise signed area between variables as a measure of confidence in the existence of a causal link. 

Many of the dynamical systems of interest can be characterized via  coupling functions, as described in \cite{stankovski2017coupling}.  In this formulation, the time evolution of the system can be represented via differential equations with functions of coupled variables. For a system of two variables this kind of system might be of the form: $$ \dot{x} = f_1(x) + g_1(x,y) \hspace{2cm}  \dot{y} = f_2(y) + g_2(x,y) $$

It is often the case however, that we do not know the true equations governing the system and are instead trying to identify the existence of causal relationships via observational time series with realizations of the process, presumably with some dynamic or measurement noise. 

\begin{figure}[H]
\centering
\includegraphics[width=1\columnwidth]{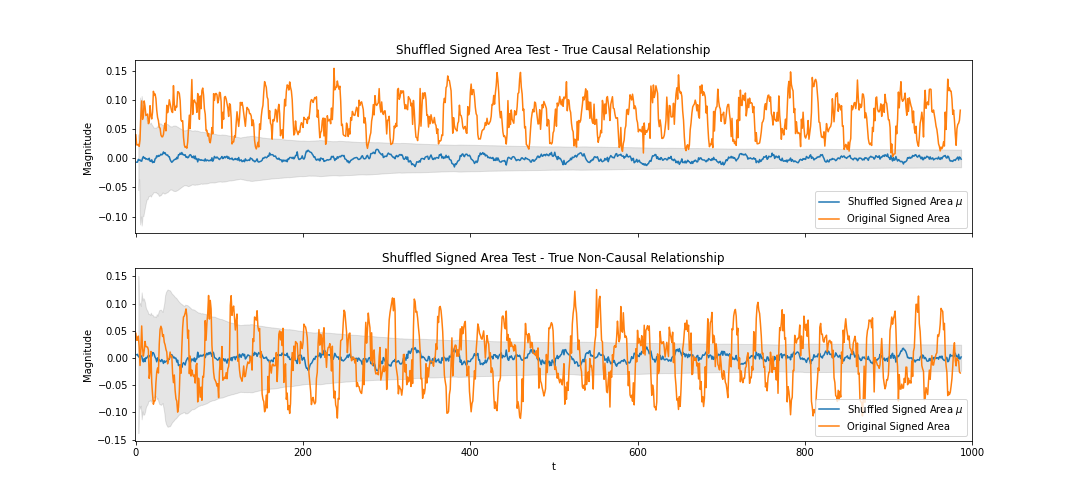}
\caption{\label{fig:shuffled_signed_area_test_example} Example comparison of shuffled signed area test for pairwise causal vs. non causal relationships. The shuffled signed area $\mu$ is the mean signed area  determined via shuffling the two time series 1000 times and computing the signed area for each pair in the ensemble. In this work, we propose a value for the confidence of a causal link existing based on the proportion of time that the actual, non-shuffled signed area remains above or below the gray shaded confidence sequence.}
\end{figure}

The primary way we identify path signatures as useful in causal discovery is via analysis of the signed area between two variables.  \cite{giusti2020iterated} proposed an approach comparing the signed area of two time series with a large ensemble of the signed areas of shuffled versions of the same time series, where they suggest that a sustained positive signed area outside of 3 standard deviations of the shuffled distribution indicates the presence of potential lag/lead relationships. We expand on this approach by recognizing that a) two variables can be causally linked without reaching this $ \pm 3 \sigma $ threshold, and thus propose a new threshold for testing the presence of lag/lead relationships based on sequential confidence sequences (see Figure ~\ref{fig:shuffled_signed_area_test_example} for a brief overview),  and b) that positive signed area alone is not enough to make conclusions about the direction of causal relationships, and thus propose that inference of causal direction in this approach should be based on the ratio of the variance between signed areas of negative and positive time-shifted versions of the time series. 

 Similar to the PCMCI algorithm with separate steps for conditional independence tests and causal direction identification, in the first step of our approach, the shuffled signed area test estimates pairwise relationships with potential causal dependencies without being capable of identifying causal direction, and the second step uses the signed area of time-shifted versions of the time series to estimate causal direction.  

The rest of the paper provides a brief review of the canonical causal discovery methods we use for baseline comparison, an introduction to path signatures and signed area, and our results building on \cite{giusti2020iterated} in using pairwise signed area to identify causal lag/lead relationships. Code for the methods described here is available at \url{https://github.com/wtglad/signed-area-causal-inference}. 

\section{Overview of Causal Discovery Methods}

There is an extensive body of literature discussing the following causal discovery algorithms, but for completeness, we will briefly review the causal discovery methods for time series that we use for baseline comparison to signature-based methods. 

\paragraph{Granger Causality}  \citet{10.2307/1912791} proposed one of the most influential statistical frameworks for testing if time series $Y$ causes time series $X$. His approach rests on two axioms - that the past and present cause the future, and that a causal variable will contain some unique information in predicting the affected variable. The general definition for discrete time series states that $Y_n$ predicts $X_{n+1}$ if $ P(X_{n+1} \in A |  \Omega_n) \neq P(X_{n+1} \in A |  \Omega_n \setminus Y_n) $ where $A$ represents some value in the sample space of $X$ and $\Omega$ represents all data about the system. 

In practice, this probability is usually established by determining a maximum time lag $ \tau $, and in the bivariate case, comparing the results of OLS regressions with and without the potentially causal variable:
$$ 1) \hspace{2.5mm} X_t = \sum_{d=1}^{\tau} \alpha_{d} X_{t-d} + \epsilon_t \hspace{2cm}  2) \hspace{2.5mm}  X_t = \sum_{d=1}^{\tau} \alpha_{d} X_{t-d} + \sum_{d=1}^{\tau} \beta_{d} Y_{t-d} + \epsilon_t $$

Whether $Y$ causes $X$ is then determined via a hypothesis test where the null hypothesis is that the $Y$ does not cause $X$. If the p-values associated with the $\beta$ coefficients (corresponding to $Y$'s explanatory input) are below a given threshold (e.g. $\alpha$ = 0.05), then we reject the null hypothesis that $Y$ does not cause $X$. 

There are a few well documented challenges with using Granger causality. For details, see \cite{chen2013regression, maziarz2015review, lusch2016inferring}, but generally, the presence of latent confounders, determininistic dynamics, nonstationarity, non-separable dynamics, and coupling, among other conditions, can result in errors in Granger causal analysis.   

For the Granger causality tests in this paper, we use the Python package \texttt{statsmodels} from \cite{seabold2010statsmodels}. 

\paragraph{Convergent Cross Mapping}

 \cite{Sugihara496} introduce convergent cross mapping (CCM) as a technique relying on state space reconstruction to determine causal relationships. Their aim is to address the nonseparability challenges associated with Granger causality and reframe the causality problem using the principle that variables in dynamical systems can be identified as causally linked if they share a common attractor manifold.
 
 This theory is based on \cite{Takens1981DetectingSA} theorem, which states that for a dynamical system whose state space is a manifold $M$ of dimension $d$,  with some function $f: M \rightarrow M $ mapping the dynamic evolution of the system, we can construct a $k$-dimensional delay embedding $\phi$ using some twice differentiable observation function $\alpha$ such that  $\phi(x) = (\alpha(x), \alpha(f(x)), ..., \alpha(f^{k-1}(x))) $. 
 
 In practice, this means that we can reconstruct a dynamical system's attractor from a single variable using its time delay embedding. The delay embeddings for $X$ and $Y$  create respective \textit{shadow manifolds} $M_X$ and $M_Y$. We can then test if there is a correspondence  between the points in $M_X$ and $M_Y$ by evaluating if points on $M_Y$ can be used to identify nearby points on $M_X$ and vice versa, as time series drawn from the same dynamical system should have similar attractor structure. 
 
There are also some challenges with this approach, such as those described in \cite{10.1371/journal.pone.0169050} - process/measurement noise and other changes to the attractor structure such as transient dynamics can complicate results. 
 
 For the CCM code in this paper, we use the Python package \texttt{skccm}, which can be found at \url{https://github.com/nickc1/skccm}. 
 
\paragraph{Causal Graphs} 

Contemporary causal inference rests on a rich body of literature developed in the last decades through techniques such as structural equation modeling, do-calculus, and probabilistic graphical models. These tools create a framework for performing causal inference about treatment effects and counterfactuals and creating probability models based on joint distributions of variables utilizing graphical structures to represent relationships between variables. For details, see works such as \cite{pearl2009causality}, \cite{Ghahramani1998}, and  \cite{Koller2009ProbabilisticGM}. 

For the purposes of this paper, we narrow our focus to the specific task of graph structure learning as a means of performing causal discovery from time series. The general task here is to construct a directed acyclic graph structure $\mathcal{G}$ comprised of nodes $X_k, k \in \{1, ..., D\}$ corresponding to $D$ time series variables that are connected by directed edges corresponding to causal relationships between variables. For many causal inference tasks, the graph structure is often assumed \textit{a priori} based on domain knowledge or otherwise determined via controlled experiments, but there are numerous techniques to attempt to recover this graph structure empirically from data. These techniques are useful when domain knowledge is lacking or controlled experiments are infeasible or unethical. 

There are too many techniques to exhaustively list them here, but one such graphical learning algorithm is the PCMCI algorithm as described in \cite{runge2020discovering}.  PCMCI works in two stages - the PC stage determines undirected edges between nodes via a series of conditional independence tests, and then the momentary conditional independence (MCI) test stage determines the causal direction between nodes. 

For true causal inference, causal graphs and in particular Bayesian networks with relationships determined by structural equations and do-calculus are likely the best tools available. The main challenges in using them arise in how to best set them up - both in articulating the correct graph structure and then specifying the underlying joint distribution feeding the probability model, particularly if this joint distribution is unknown, must be estimated from data, or is non-Gaussian. We focus on the challenge of learning structure from data here.

In this paper, we use the Python package  \texttt{Tigramite} to implement causal graph structure learning via PCMCI with partial correlation tests as described in \cite{runge2018causal}.  

\section{The Signature Method}

\paragraph{Paths and Path Signatures} 

As described in \cite{chevyrev2016primer}, the foundation of the signature method is the concept of a \textit{path}. A path $X$ in $\mathbb{R}^d$ continuously maps some interval $[a,b]$ to $\mathbb{R}^d$, which we write as $ X : [a,b] \rightarrow \mathbb{R}^d$. In the multivariate time series case, $[a,b]$ would represent some time interval, and the path $X$ would represent the time series' trajectory through  $\mathbb{R}^d$.

The \textit{path signature} is then based on iterated integrals of the variables comprising the path. The number of variables used in the iterated integral defines the \textit{depth} of the path signature. For instance, a path signature with depth 1 refers to the integration of a single variable, depth 2 refers to integration of two variables, depth 3 is three variables, etc. 

A depth 1 path signature for a variable $ i \in \{ 1, ..., d \}$ would be given by \[ S(X)_{a,t}^{i} = \int_{a<s<t}^{} dX_{s}^{i} = X_{t}^{i} - X_{a}^{i} \]

A depth 2 path signature for a pair of variables $i, j \in \{ 1, ..., d \}$ would be given by the double-iterated integral \[ S(X)_{a,t}^{i, j} = \int_{a<s<t}^{} S(X)_{a,s}^{i} dX_{s}^{j} = \int_{a<r<s<t}^{} dX_{r}^{i} dX_{s}^{j} \]

The signature can be generalized to a depth of any integer $k \geq 1 $ and collection of indices $i_1, ..., i_k \in \{ 1, ..., d \} $ to produce the \textit{k-fold iterated integral}:  \[ S(X)_{a,t}^{i_1,..., i_k} = \int_{a<s<t}^{} S(X)_{a,s}^{i_1,..., i_{k-1}} dX_{s}^{i_k} \]

We then refer to the signature of the path as the infinite collection of all the iterated integrals of $ X $: 

\[ S(X)_{a,b} = (1, S(X)_{a,b}^{1}, ..., S(X)_{a,b}^{d}, S(X)_{a,b}^{1,1}, S(X)_{a,b}^{1,2},... ) \]

In practice, however, we typically limit to some depth (e.g 2). For this research, we used the Python package \texttt{Signatory} to perform signature calculations, developed by ~\cite{signatory}. 

\paragraph{Signed Area} Also as described in  \cite{chevyrev2016primer}, we can further derive interesting information characterizing the path by computing the L\'evy area, which we will herein refer to as the \textit{signed area} associated with a depth 2 signature, defined as:   

\[A(X)_{a, b}^{i,j} := \frac{1}{2}(S(X)_{a, b}^{i,j} - S(X)_{a, b}^{j,i}) \]

Note that we can define this signed area interval $[a, b]$ over the entire path length or over shorter time intervals, and the length of this time interval is a critical hyperparameter to tune in effective causal discovery. We also note that in this paper,  we also use the notation $ A(X_i, X_j) $ to refer to the signed area $ A(X)^{i,j}$.

\section{Signed Area in Causal Identification}

\paragraph{Shuffled Signed Area Test} As discussed in \cite{giusti2020iterated}, the signed area can be used as an empirical, model-free method for identifying the presence of lag/lead relationships that may suggest the existence of causal relations between two time series. The approximate process they outline for doing so is to scale the series so that they have mean 0 and range 1, and then compute the pairwise signed area of the time series along a sliding window and compare this to the signed area for sliding windows of an ensemble of shuffled time series. Specifically, they propose shuffling the time series 1000 times and computing the pairwise signed area of a sliding window to generate a null distribution of shuffled signed area trajectories to compare with the actual trajectory. They assert that if there is a trend of positive signed area above a threshold, such as $3 \sigma$, then this suggests a potential lag/lead relationship between the two series. We will hereafter refer to this operation as the \textit{shuffled signed area test}.

We empirically observe that this $ 3 \sigma$ threshold is more extreme than necessary to return lag/lead relationships. We also observe that causal relations can present both positive and negative signed areas, so the sign alone of the signed area cannot determine causal direction. 

\paragraph{Confidence Sequence} Our alternative proposal to the $3 \sigma$ significance threshold is a confidence sequence inspired by \cite{waudbysmith2021doubly} that tightens the bounds needed for a signed area deviation from the shuffled mean to be considered significant. The act of shuffling each of the time series numerous times creates a null distribution enabling the computation of a cumulative sample mean $ \hat{\mu}_t $ and standard deviation $ \hat{\sigma}_t$ at each time point $ t $ based on the previous steps. \cite{waudbysmith2021doubly} state that for some constant $ \rho > 0 $ we can  generate a confidence sequence whose bounds are given by:  

\[ C_t = \{ \hat{\mu}_t \pm   \hat{\sigma}_t \sqrt{\frac{2 (t \rho ^ 2 + 1)}{t^2 \rho^2} log (\frac{\sqrt{t \rho^2 + 1}}{\alpha / 2})} \} \]

For simplicity in our experiments, we set $\rho = 1$, though this could certainly be further tuned. Additionally, we used the standard significance level, $\alpha = 0.05$. As $ t \rightarrow \infty $, this expression should induce converging bounds.

\paragraph{Shuffled Signed Area Deviation} More work is needed to formalize this shuffled signed area approach into a rigorous hypothesis test, but we can use these bounds, which we denote $\{C_t^-, C_t^+\}$, to assign a score associated with the significance of the signed area series' deviation from the shuffled signed area mean, which we denote the \textit{shuffled signed area deviation}, or SSAD. 

To compute SSAD, for each time index $t$, we track where the original signed area sequence $ A_t $ leaves the bounds $\{C_t^-, C_t^+\}$  and assign a value for $SSAD_t$  based on the following conditions: 
\[  SSAD_t = \left\{
        \begin{array}{ll}
            -1 & \quad A_t \leq C_t^- \\
            0 & \quad  A_t \in (C_t^-, C_t^+)  \\
            +1 & \quad A_t \geq C_t^+
        \end{array}
    \right. \] 

To compute the SSAD for the entire sequence of $T$ time steps, we evaluate: 

\[ SSAD = \frac{1}{T}\sum_{t=1}^{T} SSAD_t \]

The SSAD is then essentially a score for the proportion of time points outside the bounds, that similar to correlation, will be in the range $(-1, 1)$. For signed area sequences consistently above the upper bound of the confidence sequence, the value for SSAD will be closer to 1; for signed area sequences consistently below the lower bound, SSAD will be closer to -1; and for signed area sequences consistently within the bounds, or crucially, those that fluctuate equally above and below the bounds, SSAD will be close to 0. 

\vspace{2mm}

\begin{algorithm}[H]
\SetAlgoLined
\begin{enumerate}
\item Given $N$ time series, for each time series $X_t^i, i \in \{1, ..., N\}$, scale series so $max(X_t^i) - min(X_t^i) = 1$ and $\hat{\mu}_X = 0$.
\item Compute depth 2 signatures along sliding windows of path length $l$ for each permutation  $(X_t^i, X_t^j), i, j \in \{1, ..., N\}, i \neq j$.
\item Use depth 2 signatures to compute signed area sequences $A_t^{i, j}$ for each $i, j \in \{1, ..., N\}, i \neq j$. 
\item Create null distribution and confidence bounds for shuffled signed area:
    \begin{enumerate}
        \item Repeat steps 2. and 3. for large ensemble (e.g. 1000) of time index-shuffled versions of $X_t^i$ for all $i \in \{1, ..., N\}$
        \item Compute cumulative mean $\hat{\mu}_t$ and sample variance $\hat{\sigma}_t$ for each of the time index-shuffled signed areas. 
        \item Use $\hat{\mu}_t$, $\hat{\sigma}_t$ to generate confidence sequence bounds $\{C_t^-, C_t^+\}$ for each pair $i, j \in \{1, ..., N\}, i \neq j$. 
    \end{enumerate}
\item Compute $SSAD_t$ for each time step of $A_t^{i, j}$ based on $\{C_t^-, C_t^+\}$ and compute SSAD score for entire sequence. 
\end{enumerate}
 \caption{Causal Dependence via Shuffled Signed Area Test with Confidence Sequence}
\end{algorithm}

\paragraph{Time-Shifted Signed Area Variance Ratio Test for Causal Direction}

Because the signed area alone is only capable of suggesting a causal link between variables without suggesting direction of those causal relations, we need another test in order to identify that direction. We introduce the \textit{time-shifted signed area variance ratio test} (TS-SAVR) as a method to leverage differences in  signed area variance over past and future time lags in order to infer causal direction. 
\vspace{1mm}    

\begin{algorithm}[H]
\SetAlgoLined
\begin{enumerate}
    \item For time shifts $\tau $ in some interval of negative to positive time shifts $[\tau_{min}, \tau_{max}]$, create shifted versions of each variable $X_{t+\tau}^i$ for $i \in \{1, ..., N\}$. For our experiments, we used $ \tau \in [-10, 10]$.
    \item For each time-shifted variable $i \in \{1, ..., N\} $: 
        \begin{enumerate}
            \item Compute pairwise signed area $ A(X_{t+\tau}^i, X_t^j)$ for each $\tau \in [\tau_{min}, \tau_{max}]$ and variable $j$ in the set of non time-shifted variables  $\{1, ..., N\} \setminus i$ over the entire time interval.
            \item Find the two variances of signed area values for $ \tau < 0$ and $ \tau > 0$, which we denote $Var(A_{\tau^-}), Var(A_{\tau^+})$ . Note that we exclude $A_{\tau = 0}$.
            \item Compute time-shifted signed area variance ratio $ Var(A_{\tau^-})/Var(A_{\tau^+}) $:
            \begin{itemize}
                \item If  $Var(A_{\tau^-})/Var(A_{\tau^+})  \geq 1.1$, this suggests causal direction from $i \rightarrow j$
                \item If $Var(A_{\tau^-})/Var(A_{\tau^+})  \leq 0.9$, this suggests $j \rightarrow i$.
                \item If $Var(A_{\tau^-})/Var(A_{\tau^+})  \in (0.9, 1.1) $, this suggests mutual causation, $ i \leftrightarrow j $. \end{itemize}.  
        \end{enumerate}
\end{enumerate}
 \caption{Time-Shifted Signed Area Ratio Test for Causal Direction}
\end{algorithm}

\paragraph{Signed Area-Based Causal Discovery} Tying our two algorithms together, we first identify candidate pairs of time series exhibiting causal dependence via the shuffled signed area test, and then for each of these pairs, determine their causal direction via the time-shifted signed area ratio test. Based on the results of this test, we can then take the absolute value of the SSAD to return the confidence of a causal link existing. We can then optionally choose some threshold (e.g 0.5) over which we say a causal link exists to create an edge in a proposed causal graph $\cal{G}$ or simply use $|SSAD|$ as is. We do observe empirically that 0.5 may be too high a threshold, so for our results, we use the raw $|SSAD|$ and interpret our results as a relative score.

\section{Results}

We explore the application of the shuffled signed area test  and the associated time-shifted signed area ratio test causal discovery method on a number of both synthetic and real world time series datasets, with baseline comparisons to other canonical causal discovery methods - namely Granger causality, convergent cross mapping, and PCMCI. 

 As some low dimensional synthetic and real world examples to compare with canonical methods, we examine the following systems from \cite{ye2015distinguishing}.  Because these systems only contain causally linked variables, in order to determine the method's ability to distinguish causal from non-causal relationships, we add a white noise variable $W_t$ to each analysis, where the entire series $W_t \sim \mathcal{N}(0,\,1) $ is the same length as the raw time series and subject to the same scaling method as the raw time series. We then evaluate how well signed area-based causal discovery performs in returning the true causal links. 

As baseline comparisons, for each pair of features within each system, we show the Granger causality test's minimum p-value from the sum of squared residuals (SSR) f-test, the maximum convergent cross mapping skill measured by $R^2$, and the minimum p-value returned from PCMCI conducted with a partial correlation independence test. For the Granger causality and PCMCI tests we use time lags $ \tau \in [-10, 0] $ and use the minimum p-value across these lags and a significance threshold of $\alpha = 0.05$ in order to produce a binary estimate whether a directed causal link exists from $X_i \rightarrow X_j$, without attempting to identify specific temporal links. For the convergent cross mapping test, we take the maximum $R^2$ value corresponding to cross map skill and compare these maxima for the two variables - if one score is close to 1 and substantially higher than the other, then CCM suggests this variable is causing the other. 

\begin{enumerate}

    \item \textbf{Two-Species Model System with Synchrony} 

As the simplest example of a low dimensional system with a unidirectional causal relation, consider the system

    \[ X_{t+1} = X_t (3.8 - 3.8 X_t) \]
    \[ Y_{t+1} = Y_t (3.1 - 3.1 Y_t - 0.8 X_t) \]

where system is initialized with $X_0 = 0.2, Y_0 = 0.4$ and run for 1000 time steps. 

 Figure ~\ref{fig:2spec_synch_plots} shows a subset of feature pairs' shuffled signed area results plotting the actual signed area series vs. the shuffled distribution using a path length of 10 time steps, as well as the time-shifted signed areas used to compute the variance ratio. Table  ~\ref{tab:2spec_synch_table} shows the numeric values associated with each of these tests for pairwise combinations of all three features as well as the baseline comparisons described at the beginning of this section. Feature pairs marked with an asterisk (e.g. $A(X, Y)*$ ) indicate causal links, and baseline statistics marked with a double asterisk (e.g. $0.05**$) suggest the presence of a causal relationship.

\begin{figure}[H]
\centering
\includegraphics[width=1\columnwidth]{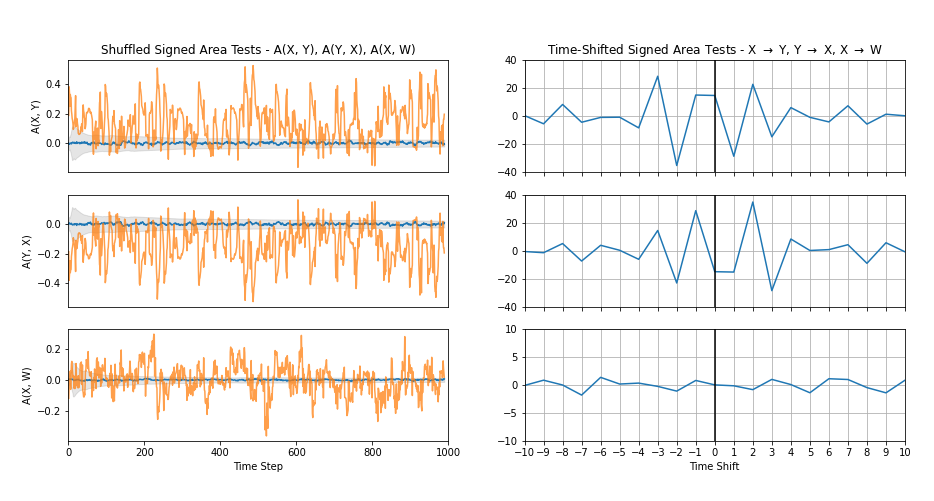}
\caption{Results of shuffled signed area test and time-shifted signed areas. The shuffled signed area tests demonstrate that $A(X, Y)$ is consistently above the confidence sequence bounds, $A(Y, X) = - A(X, Y) $ is consistently below the sequence bounds, and $ A(X, W) $ alternates above and below the bounds. The corresponding time-shifted signed areas show a negative to positive variance ratio $ \geq 1.1 $ for $ X \rightarrow Y$, correctly suggesting this causal relationship in this system.  }
\label{fig:2spec_synch_plots} 
\end{figure}

\begin{table}[H]
\centering
\begin{tabular} {  p{0.12\textwidth}  p{0.12\textwidth}  p{0.12\textwidth}  p{0.10\textwidth}  p{0.10\textwidth}  p{0.12\textwidth}  } 
\hline
\bfseries Feature Pair & \bfseries SSAD & \bfseries TS-SAVR & 
\bfseries GC min p-val & \bfseries CCM max $R^2$ & \bfseries PCMCI min p-val  \\
\hline
 (X, Y)* & 0.7098 & 1.4776  & 0.0000** & 0.9984** & 0.0000**\\
\hline
(Y, X) & -0.7098 & 0.6813  & 0.0000** & 0.8525 & 0.0000**  \\
\hline
(Y, W) & 0.1355 & 2.5740  & 0.2813 & 0.0862 & 0.0203**  \\
\hline
(W, Y)  & -0.1355 & 0.4586  & 0.1348 & 0.0772 & 0.2457  \\
\hline
(X, W) & 0.0172 & 0.8997  & 0.4771 & 0.1030 & 0.0145**  \\
\hline
(W, X) & -0.0172 & 1.1704  & 0.3489 & 0.0903 &  0.1048  \\
\hline
\end{tabular}
\caption{Numeric results for shuffled signed area deviation (SSAD), time-shifted signed area variance ratio (TS-SAVR), and comparison to baseline methods. Note that Granger causality and PCMCI appear to be suggesting an incorrect causal link from $ Y \rightarrow X$, and PCMCI also returns some false dependencies from $ X\rightarrow W $ and $ Y \rightarrow W$.  }
\label{tab:2spec_synch_table}
\end{table}

    \item \textbf{Two-Species Model System with Bidirectional Causality} 
    \[ X_{t+1} = X_t (3.78 - 3.78 X_t - 0.07 Y_t) \]
    \[ Y_{t+1} = Y_t (3.77 - 3.77 Y_t - 0.08 X_{t-\tau_d}) \]
where $\tau_d \in \{0, 2, 4\}$ and represents the time delay of $X_t$'s impact on $Y_t$. The system is initialized with $X_0 = 0.2, Y_0 = 0.4$ and run for 3000 time steps.  For the $\tau_d \in \{2, 4\}$ cases we repeat the same initial conditions to enable the time lags.     

Table ~\ref{tab:2spec_bidirec_tau0_table} shows the results of the signed area causal discovery algorithms for the $\tau_d = 0$ case with no time delay.  For brevity, we do not include the results from $ \tau \in \{2, 4\}$ here as we observe similar behavior to the $\tau_d = 0$ case - namely, that this method struggles with bidirectional causal relationships even with the presence of a time lag.  

\begin{table}[H]
\centering
\begin{tabular} {  p{0.12\textwidth}  p{0.12\textwidth}  p{0.12\textwidth}  p{0.10\textwidth}  p{0.10\textwidth}  p{0.12\textwidth}  } 
\hline
\bfseries Feature Pair & \bfseries SSAD & \bfseries TS-SAVR & 
\bfseries GC min p-val & \bfseries CCM max $R^2$ & \bfseries PCMCI min p-val  \\
\hline
 (X, Y)* & -0.0860 & 0.9512  & 0.0000** & 0.9509** & 0.0000**\\
\hline
(Y, X)* & 0.0860 & 1.0803  & 0.0000** & 0.9454** & 0.0000**  \\
\hline
(Y, W) & -0.0097 & 0.8118  & 0.0252 & 0.0516 & 0.0901  \\
\hline
(W, Y)  & 0.0097 & 1.1820  & 0.0961 & 0.0143 & 0.0343**  \\
\hline
(X, W) & -0.0094 & 4.0825  & 0.3260 & 0.0365 & 0.0451**  \\
\hline
(W, X) & 0.0094 & 0.2652  & 0.3886 & 0.0529 &  0.1134  \\
\hline
\end{tabular}
\caption{Signed area causal discovery results with comparison to baseline methods. We note that the SSAD does not reach a 0.5 threshold so by this strict definition, the method does not recover the bidirectional causal relationship with no time lag, though the SSAD is an order of magnitude higher than the variable-noise pairs' scores. We also note that in the case of bidirectional causality, the TS-SAVR is much closer to 1 than the case of unidirectional causality. We note also that the baseline methods appear to be capturing the bidirectional causality but PCMCI has some false positives returned for $ W\rightarrow Y$ and $ X\rightarrow W$. }
\label{tab:2spec_bidirec_tau0_table}
\end{table}

    \item \textbf{Four-Species Model System} 
    
    \[ V_{t+1} = V_t (3.9 - 3.9 V_t) \]
    \[ X_{t+1} = X_t (3.6 - 0.4 V_t - 3.6 X_t) \]
    \[ Y_{t+1} = Y_t (3.6 - 0.4 X_t - 3.6 Y_t) \]
    \[ Z_{t+1} = Z_t (3.8 - 0.35 Y_t - 3.8 Z_t) \]
    
The system is initialized with $V_0 = X_0 = Y_0 = Z_0 = 0.4$ and run for 1000 time steps. Table ~\ref{tab:4spec_table} captures the numeric results of signed area causal discovery. 

\begin{table}[H]
\centering
\begin{tabular} {  p{0.12\textwidth}  p{0.12\textwidth}  p{0.12\textwidth}  p{0.10\textwidth}  p{0.10\textwidth}  p{0.12\textwidth}  } 
\hline
\bfseries Feature Pair & \bfseries SSAD & \bfseries TS-SAVR & 
\bfseries GC min p-val & \bfseries CCM max $R^2$ & \bfseries PCMCI min p-val  \\ \hline
(V, X)* & -0.8868 & 12.7940  & 0.0000** & 0.9841** & 0.0000** \\ \hline
(X, V) & 0.8868 & 0.0818  & 0.0000** & 0.3425 & 0.0964 \\ \hline
(X, Y)* & -0.7422 & 3.2356  & 0.0000** & 0.9327** & 0.0000** \\ \hline
(Y, X) & 0.7422 & 0.3000  & 0.0349** & 0.1982 & 0.0286** \\ \hline
(Z, X) & 0.1618 & 0.1638  & 0.0802 & 0.1509 & 0.0078** \\ \hline
(X, Z) & -0.1618 & 6.3977  & 0.0000** & 0.2282 & 0.0078** \\ \hline
(V, Y) & 0.1405 & 16.8616  & 0.0000** & 0.2547 & 0.0012** \\ \hline
(Y, V) & -0.1405 & 0.0608  & 0.2105 & 0.1570 & 0.2246 \\ \hline
(W, X) & -0.1203 & 1.4477  & 0.0589 & 0.0588 & 0.0638 \\ \hline
(X, W) & 0.1203 & 0.6206  & 0.3071 & 0.0863 & 0.1140 \\ \hline
(Y, Z)* & -0.1183 & 3.6474  & 0.0000** & 0.9571** & 0.0000** \\ \hline
(Z, Y) & 0.1183 & 0.2578  & 0.2124 & 0.2472 & 0.0100** \\ \hline
(V, Z) & 0.0910 & 3.9451  & 0.0113** & 0.0301 & 0.0379** \\ \hline
(Z, V) & -0.0910 & 0.3207  & 0.0562 & 0.1007 & 0.0590 \\ \hline
(W, Y) & 0.0506 & 1.8810  & 0.1955 & 0.0526 & 0.0061** \\ \hline
(Y, W) & -0.0506 & 0.4770  & 0.2513 & 0.0939 & 0.1217 \\ \hline
(V, W) & -0.0425 & 1.0104  & 0.0530 & 0.0804 & 0.0053** \\ \hline
(W, V) & 0.0425 & 1.0311  & 0.2559 & 0.1063 & 0.0053** \\ \hline
(W, Z) & -0.0222 & 2.2167  & 0.0905 & 0.0699 & 0.0153** \\ \hline
(Z, W) & 0.0222 & 0.4292  & 0.0153** & 0.0728 & 0.0110** \\ \hline

\hline
\end{tabular}
\end{table}
\begin{table}[H]
\caption{Signed area causal discovery results with comparison to baseline methods for the 4 species system. Here we note that the 0.5 threshold fails to capture a true causal relationship from $ Y \rightarrow Z $, but we note that if viewed as an absolute ranking with a lower threshold instead of using 0.5, we would be able to recover this relationship. We also note that if this lower threshold is used, we do include false  positives from $V \rightarrow Y$, $ X \rightarrow Z$, and $W \rightarrow X$. The $W \rightarrow X$ is indeed an erroneous false positive, but the other two connections capture an important point about this method - namely that for hierarchical causal relationships with non-Markovian dependencies, there can still be a lag-lead relationship. We also note that Granger causality and PCMCI also struggle with false positives with non-Markovian dependencies.  }
\label{tab:4spec_table}
\end{table}

    \item \textbf{Paramecium-Didinium Predator-Prey System}
    
This dataset captures a classic predator-prey system based on experimental time series of protozoa populations from \cite{10.2307/4195}. Data are taken from figure 11a, tabulated in \url{https://robjhyndman.com/tsdldata/data/veilleux.dat}. Here we denote paramecium population as $X$ and the didinium population as $Y$ and assume that like the standard predator-prey model, there should be bidirectional causality $ X \leftrightarrow Y$. Table ~\ref{tab:paramecium_didinium_table} captures the results of signed area causal discovery.

\begin{table}[H]
\centering
\begin{tabular} { p{0.12\textwidth}  p{0.12\textwidth}  p{0.12\textwidth}  p{0.10\textwidth}  p{0.10\textwidth}  p{0.12\textwidth}  } 
\hline
\bfseries Feature Pair & \bfseries SSAD & \bfseries TS-SAVR & 
\bfseries GC min p-val & \bfseries CCM max $R^2$ & \bfseries PCMCI min p-val  \\
\hline
(X, Y)* & 1.0000 & 0.9400  & 0.0000** & 0.9201** & 0.0077** \\ \hline
(Y, X)* & -1.0000 & 0.5999  & 0.0000** & 0.8729** & 0.1696 \\ \hline
(Y, W) & 0.1167 & 0.7774  & 0.1606 & 0.0351 & 0.0373** \\ \hline
(W, Y) & -0.1167 & 1.2766  & 0.1589 & 0.1401 & 0.0138** \\ \hline
(X, W) & -0.0333 & 1.6437  & 0.2319 & 0.0872 & 0.3815 \\ \hline
(W, X) & 0.0333 & 0.6186  & 0.0779 & 0.1694 & 0.0178** \\ \hline
\end{tabular}
\caption{Signed area causal discovery results with comparison to baseline methods for Paramecium-Didinium dataset. We note that the small sample size ($n=71$) drives the extreme values of $\pm 1$ for SSAD and that TS-SAVR suggests bidirectional causality based on the $ X \rightarrow Y$ test but not for $Y \rightarrow X$. Granger causality and CCM both appear to be able to recover the bidirectional causality, and PCMCI only returns $ X\rightarrow Y$ and several false positives.  }
\label{tab:paramecium_didinium_table}
\end{table}

    \item \textbf{Vostok Ice Core \ch{CO2} and Temperature}
    
This dataset captures ice core measurements tabulating atmospheric carbon dioxide concentration and temperature variation for 420,000 years of data as discussed in \cite{article}. Data are tabulated at \url{https://cdiac.ess-dive.lbl.gov/trends/co2/vostok.html}. As done in \cite{ye2015distinguishing}, we linearly interpolate the data to have evenly sampled measurements every 1000 years.  Here we denote \ch{CO2} as $X$ and temperature variation as $Y$ and note that the primary causal relation we should see is $ X \rightarrow Y$ due to the greenhouse effect. 

\begin{table}[H]
\centering
\begin{tabular} {  p{0.12\textwidth}  p{0.12\textwidth}  p{0.12\textwidth}  p{0.10\textwidth}  p{0.10\textwidth}  p{0.12\textwidth}  } 
\hline
\bfseries Feature Pair & \bfseries SSAD & \bfseries TS-SAVR & 
\bfseries GC min p-val & \bfseries CCM max $R^2$ & \bfseries PCMCI min p-val  \\
\hline
(X, Y)* & 0.1141 & 76.2777  & 0.0000** & 0.7213** & 0.0246** \\ \hline
(Y, X) & -0.1141 & 0.0120  & 0.0121** & 0.6292 & 0.0509 \\ \hline
(Y, W) & -0.0372 & 1.7632  & 0.0859 & 0.0699 & 0.0752 \\ \hline
(W, Y) & 0.0372 & 0.4381  & 0.6261 & 0.0790 & 0.1831 \\ \hline
(X, W) & 0.0174 & 5.5722  & 0.1941 & 0.1233 & 0.0093** \\ \hline
(W, X) & -0.0174 & 0.5988  & 0.8615 & 0.0495 & 0.0737 \\ \hline
\hline
\end{tabular}
\caption{Signed area causal discovery results with comparison to baseline methods for Vostok ice core. We note that while the $ X \rightarrow Y$ is not returned at the $|SSAD| > 0.5$ level, it is still an order of magnitude higher than the other SSAD values and the TS-SAVR strongly suggests the direction is $ X \rightarrow Y$. We also note that Granger causality and PCMCI, while able to significantly recover $ X \rightarrow Y$, also return other false positives. }
\label{tab:vostok_table}
\end{table}

\end{enumerate}

\section{Discussion and Conclusion}

Based on our results, we conclude that most appropriate initial application of this signed area-based method appears to be causal discovery in low dimensional, coupled time series with unidirectional causal relationships. We believe the chief contribution here is a totally model-free, data-driven approach to uncovering causal relationships in coupled time series that provides a useful alternative to traditional methods such as Granger causality, convergent cross mapping, and causal graph discovery algorithms such as PCMCI. 

In this sense, this method is likely most applicable in the initial stages of exploratory data analysis on potentially coupled time series in order to identify candidates for causal links and inform an initial construction of causal graph structures. In order to perform rigorous causal analysis, controlled experiments or probabilistic modeling will still be required. However, the full probabilistic model specification can be computationally expensive especially for high dimensional data and for data whose joint distributions are non-Gaussian. In these cases, the shuffled signed area test with a confidence sequence may help more rapidly identify candidates for causal graph structures that can then lean on more traditional methods for causal graph learning. 

There are a number of areas for further research and refinement to make this a more robust method that we have not explored here. As shown, this approach struggles with bidirectional causation, and is not able to distinguish when variables have non-Markovian dependencies or shared confounders vs. causing one another directly. At present, its power is thus in its ability to detect empirical lag/lead relationships, not in estimating causal treatment effects or determining hierarchical causal graph structures. 

We note that the empirical results indicate that using a basic $|SSAD|$ value as the probability of a causal link needs additional calibration, as we can have true causal links below the 0.5 threshold, and we should thus perhaps view this as more of a relative, not absolute, score. The potential pitfall of totally relying on a relative approach would be spurious connections identified in a system where there are no true causal relations. Further work is thus needed to develop the reliability of the bounds of confidence sequence and the associated sequential hypothesis and shuffled signed area tests in order to reliably embed this SSAD score into a well-calibrated probability measure. 

We also note that more work is needed to understand the properties of the sliding window or signature path length hyperparameter in applying this method. We observe that if the path length is too low or high compared to the time lag between causal variables, it may fail to capture relevant causal links or otherwise produce spurious links. In this work we maintained consistent path lengths across examples, and as an immediate next step it would be useful to understand bifurcation points where causal relationships can and cannot be returned based on varying this path length. We suspect for periodic systems or systems with a fixed time dependencies the optimal path length with be some function of these characteristics, but further research is needed to explore this fully. 

A natural extension of this work would be to explore what kinds of systems and assumptions are required in order for this method to be effective. For instance, in the Vostok data, we observe a high degree of non-stationarity and extreme values within the data that may be impacting the results even when differencing the data. We also wonder what assumptions about nonlinearity, differentiability, etc. are required in order to reliably perform causal inference this way. 

Finally, it would also be useful to explore other properties of path signatures in carrying out causal analysis. We hypothesize that first order signature terms should embed useful historical, aggregated information about time series that should enhance study of certain systems. It would also be valuable to assess if higher order (depth 3+) path signatures are able to inform any useful causal analysis. 

To conclude, via the signed area-based approaches we have outlined here, we believe we offer a novel, flexible approach to data-driven, model free causal discovery that performs comparably to existing methods for identifying causal relationships.

\medskip

\bibliography{references}

\end{document}